\documentclass[conference]{IEEEtran}
\usepackage[nocompress]{cite}
\usepackage[pdftex]{graphicx}
\usepackage{amsmath}
\usepackage{float}
\usepackage{array}
\usepackage{url}
\usepackage{boldline}
\usepackage{subfig}

\makeatletter
\def\ps@IEEEtitlepagestyle{%
  \def\@oddfoot{\mycopyrightnotice}%
  \def\@evenfoot{}%
}
\def\mycopyrightnotice{
{ \footnotesize
  \begin{minipage}{\textwidth}
  \centering
  \copyright2016 IEEE. Personal use of this material is permitted. Permission from IEEE must be obtained for all other uses, in any current or future media, including reprinting /republishing this material for advertising or promotional purposes, creating new collective works, for resale or redistribution to servers or lists, or reuse of any copyrighted component of this work in other works. \hfill
\end{minipage}
}
}

\begin{document}
\title{Dataset Augmentation for Pose and Lighting Invariant Face Recognition}
\author{\IEEEauthorblockN{Daniel Crispell\IEEEauthorrefmark{1}, Octavian Biris\IEEEauthorrefmark{1}, Nate Crosswhite\IEEEauthorrefmark{2}, Jeffrey Byrne\IEEEauthorrefmark{2}, Joseph L. Mundy\IEEEauthorrefmark{1}}
  \IEEEauthorblockA{\IEEEauthorrefmark{1}Vision Systems, Inc.}
  \IEEEauthorblockA{\IEEEauthorrefmark{2}Systems and Technology Research}
}


\maketitle

\begin{abstract}
The performance of modern face recognition systems is a function of the dataset on which they are trained.
Most datasets are largely biased toward ``near-frontal'' views with benign lighting conditions, negatively effecting recognition performance on images that do not meet these criteria.
The proposed approach demonstrates how a baseline training set can be augmented to increase pose and lighting variability using semi-synthetic images with simulated pose and lighting conditions.
The semi-synthetic images are generated using a fast and robust 3-d shape estimation and rendering pipeline which includes the full head and background.
Various methods of incorporating the semi-synthetic renderings into the training procedure of a state of the art deep neural network-based recognition system without modifying the structure of the network itself are investigated.
Quantitative results are presented on the challenging IJB-A identification dataset using a state of the art recognition pipeline as a baseline.
\end{abstract}

\section{Introduction}
Performance of automatic face recognition algorithms has increased dramatically over the course of the past few years, primarily due to a combination of the proliferation of algorithms based on deep convolutional neural networks and the availability of large labeled face datasets.
While these algorithms show promise in the direction of reliable face identification in completely unconstrained pose, lighting, and expression (PIE) conditions, the degree to which performance is invariant under such conditions is largely a function of the degree to which the conditions are present in the training data.
On the other hand, collecting a large (greater than one million images) labeled face dataset that densely covers the full range of conditions expected to be encountered in a given test set is often impossible or impractical.
Such datasets are typically gathered from images available on the web, a large percentage of which are taken by professional photographers with the primary objective of capturing the face of the imaged subject.
As such, the lighting conditions are typically very benign, and the subject's face is often directly facing the camera.
In order to combat this bias, a training data augmentation pipeline is proposed in which the 3-d shape of each face in a baseline training set is estimated, enabling the face to be rendered with arbitrarily imposed pose and lighting.

\subsection{Contributions}
The contributions of this work are:
\begin{itemize}
  \item A fast, robust 3-d face shape estimation algorithm capable of operating on single or multiple images.
  \item An efficient rendering pipeline capable of arbitrarily reposing faces given their 3-d reconstruction.
  \item An efficient face relighting algorithm.
  \item Quantitative investigation of the effect of training set pose and lighting augmentation on face recognition results.
\end{itemize}

\section{Related Work}
3-d reconstruction of face shape from 2-d imagery has long been an active area of research.
Blanz and Vetter~\cite{blanz_99} introduced their ``3-D Morphable Model'', which represented the space of human face shapes using principal component analysis (PCA) of mesh vertex locations.
More recently, Roth et al.~\cite{roth_15} introduced a reconstruction algorithm that uses 2-d facial landmarks as constraints on a warping of a 3-d mesh, followed by a ``fine tuning'' of the 3-d shape based on photometric stereo.
Zhu et al.~\cite{zhu_15} also use 2-d facial landmarks to estimate face shape, but incorporate a variation on Blanz and Vetter's 3-d Morphable Model~\cite{blanz_99} that includes expression parameters as a shape prior to prevent unnatural or improbable shapes from being reconstructed.
In order to deal with the viewpoint-dependence of the jawline landmarks, the constraints on their respective positions is relaxed.
The proposed approach is similar in that it uses 2-d landmarks as constraints on a 3-d Morphable Model that includes expression parameters, but it deals with viewpoint-dependent landmarks uniformly using an iterative estimate/render/detect procedure rather than relaxing constraints on specific landmark positions.

Given the capability to estimate a 3-d face model, there exist several approaches for leveraging such models for the purpose of improved facial recognition performance.
Perhaps the most straightforward of such approaches is to use 3-d measurements of the face directly for recognition.
While this approach has been shown feasible given high enough reconstruction fidelity~\cite{gupta_10, paysan_09}, such precision is generally not feasible to extract from a single (potentially low resolution) image in an unconstrained environment.

Rather than recognize directly in 3-d, a second class of approaches instead leverages an estimated 3-d model for the purpose of removing unwanted nuisance variables such as pose and expression from the test images.
Taigman et al.~\cite{taigman_13} demonstrate that transforming facial imagery such that the subject appears to be directly facing the camera (i.e.\ ``frontalization'') gives measurable performance gains, leading to state of the art performance at the time.
Chu et al.~\cite{chu_14} perform expression normalization in addition to pose normalization.
While these approaches are capable of removing small pose variation from a test image, they are unable to handle extreme transformations such conversion from a profile view to a frontal view.
For this reason, the approach alone is not able to compensate for a completely unconstrained environment.

Rather than using normalization to make the recognition task easier at test time, a third approach is to instead render challenging views to make the recognition task more challenging at training time.
In theory, this method forces the network to learn features which are invariant to the transformations included in the training set.
Masi et al.~\cite{masi_16} demonstrate the feasibility of such an approach by augmenting a baseline training set with renderings at fixed yaw angle intervals.
Rather than estimating 3-d shape in the original image, renderings are produced using a set of ``exemplar'' face shapes for each subject.
The proposed approach is similar in goal but has several important distinctions.
The 3-d shape of the subject in a given image is estimated before rendering in order to simulate viewpoint change as accurately as possible.
Rather than rendering the training subjects from a small, fixed number of views, viewpoint in yaw, pitch, and roll is randomly perturbed to achieve maximal pose invariance.
We refer to this random perturbation as ``pose jittering''.
Finally, a rendering procedure for altering lighting conditions in addition to pose is presented (``lighting jittering'') in the proposed approach.

\section{Shape and Camera Estimation}
\label{sec:shape_and_camera_est}

In order to properly render novel views of faces in input imagery, both the 3-d geometry of the observed head and the projection model of the observing camera must be estimated.
The projection model is assumed to be orthographic, so six parameters are needed:  A scaling constant $s$ that converts world units (mm in this case) to pixels, a 2-d image offset $t_u, t_v$, and the three pose parameters (yaw,pitch,roll) that parameterize the rotation matrix $R$.

\begin{equation}
  \label{eq:orthographic_projection}
  \begin{bmatrix}
    u \\
    v
  \end{bmatrix}
  =
  \begin{bmatrix}
    s & 0 & 0 \\
    0 & s & 0 \\
  \end{bmatrix}
  R x +
  \begin{bmatrix}
    t_u \\
    t_v
  \end{bmatrix}
\end{equation}

\begin{figure}[h!]
  \includegraphics[width=0.5\textwidth]{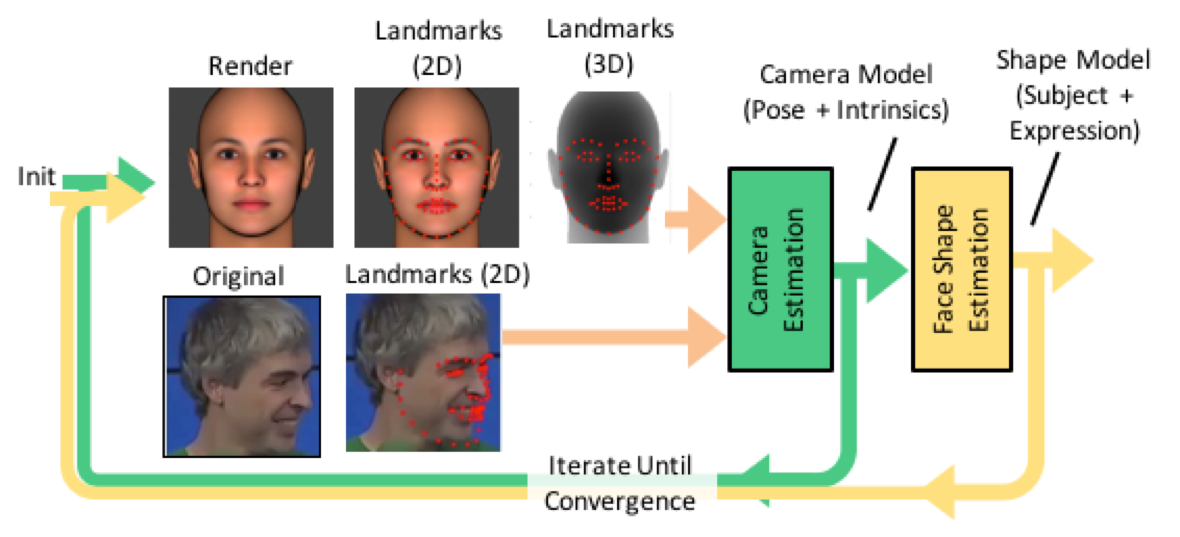}
  \caption{A flowchart of the camera, pose, and face shape estimation pipeline.}
\label{fig:estimation_flow}
\end{figure}

The orthographic projection assumption generally holds in the neighborhood of the target face for most imagery captured at significant distance from the subject, but may break down for close-range images such as those captured by a laptop webcam.
The world coordinate system is tied to the 3-d head model, so estimating the camera pose is equivalent to estimating the pose of the head.
The basic estimation pipeline is shown in Figure~\ref{fig:estimation_flow}.
The estimation is framed as an optimization procedure in which a rendered synthetic head model is aligned with the input face image.
The location of a set of image landmarks is used to measure the quality of alignment.
A cascaded regression tree algorithm~\cite{kazemi_14} as implemented in the Dlib software library~\cite{dlib_09} is used to detect the landmark locations in both the input image and the rendered synthetic image.

Once detected, the landmarks corresponding to the synthetic rendering are backprojected onto the underlying 3-d model in order to obtain their 3-d positions.
The correspondences of the 2-d landmark locations in the input image and the 3-d locations on the model are then used to estimate an affine projection matrix using the direct linear transform (DLT)~\cite{hartley_zisserman}.
The $RQ$ matrix decomposition algorithm is used to decompose the affine projection matrix into the product of an upper triangular projection matrix and a rotation.
The projection matrix is further decomposed into an orthographic projection matrix (square pixels, zero skew) and a residual affine transformation which is typically small and discarded.
The pose of the head relative to the camera is described by the rotation matrix, and the internal parameters of the camera (scale and 2-d offset) are described by the orthographic projection matrix.

The shape of the head is described by the location of $N$ 3-d mesh vertex positions with fixed topology.
A set of synthetic training data (1000 randomly generated head models with neutral expression) is generated using the FaceGen~\cite{facegen} software application (the ``subject'' dataset).
A smaller set of 3-d models are generated exhibiting various facial expressions (the ``expression'' dataset).
Principal component analysis (PCA) is then used to generate a linear transformation that mapped the vertex positions into a new space such that each dimension spans the direction of maximal variance, with the constraint that it is orthogonal to all previous dimensions.
This procedure is performed independently for the ``subject'' and ``expression'' datasets, resulting in two sets of PCA basis vectors, $A$ and $B$, respectively.
Similar to Chu et al.~\cite{chu_14} and Zhu et al.~\cite{zhu_15}, the shape variations due to expression are assumed to be independent to those intrinsic to the subject, and the full shape variation is described according to the composition of the linear PCA models.

\begin{equation}
  \label{eq:PCA}
  V = \bar{V} + \alpha A + \beta B
\end{equation}

The coefficient vectors $\alpha$ and $\beta$ describe the subject's intrinsic face shape and shape variation due to expression, respectively.

\begin{figure*}[ht]
  \includegraphics[width=1.0\textwidth]{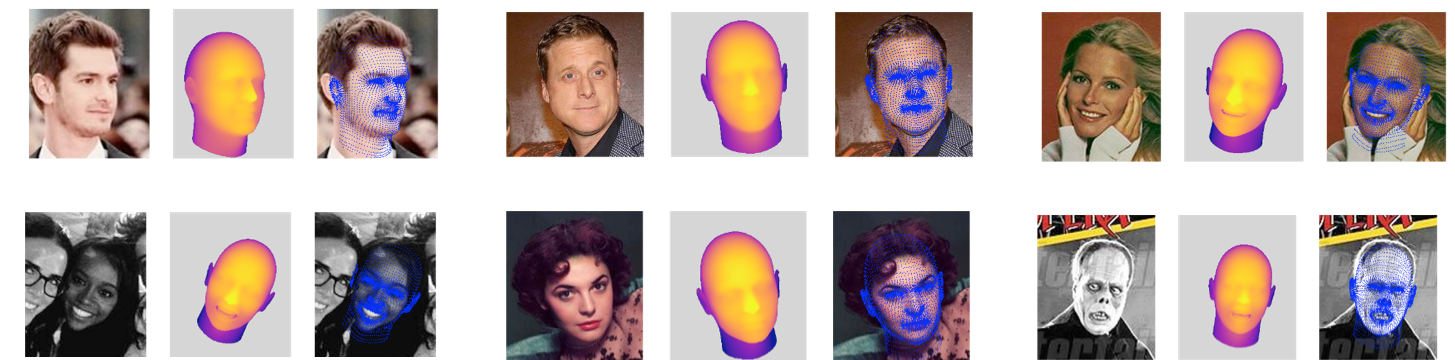}
  \caption{Visualization of statistical model fitting results on samples from the training imagery.
Note that although the algorithm is capable of jointly estimating shape given multiple images, the results shown are generated using a single image only.
From left to right: The input image, a depth map of the optimized geometry and pose, and the projection of the optimized mesh vertices onto the input image to visualize alignment accuracy.
}
\label{fig:vgg-face_estimated}
\end{figure*}

\subsection{Optimization}
Using the principal component vectors, any set of 3-d vertex coordinates with the same topology as the input data can be transformed to coefficients $\alpha$ and $\beta$.
In order to compute the coefficient values from image observations, the PCA basis vectors must be projected into the image plane using estimated projection matrices of the input images.
Additionally, the only known image coordinates are those of the detected landmarks.
The elements of the ``subject'' and ``expression'' PCA vectors corresponding to mesh vertices nearest to the backprojected landmarks are extracted (matrices $A$ and $B$, respectively) and multiplied by the estimated camera projection matrices ($P_n$ for image $n$) to project into the image plane.
The values of the subject and expression PCA coefficients ($\alpha$ and $\beta$, respectively) are then solved for using singular value decomposition (SVD).
Multiple observations can easily be incorporated using the same framework, and a single set of ``subject'' coefficients plus one set of ``expression'' coefficients per image are solved for as shown in Equation~\ref{eq:stat_model_solve}, where the difference between the projected landmarks and the expected position of the corresponding landmarks on the ``mean head'' geometry in image $n$ is denoted $\Delta s_n$.

\begin{equation}
\label{eq:stat_model_solve}
  \begin{bmatrix}
    P_0 A & P_0 B  & 0      & \ldots   \\
  \vdots     & 0            & \ddots &          \\
  P_N A & 0            &        & P_N B
\end{bmatrix}
\begin{bmatrix}
  \alpha    \\
  \beta_0 \\
  \vdots         \\
  \beta_N
\end{bmatrix}
=
\begin{bmatrix}
  \Delta s_0 \\
  \vdots     \\
  \Delta s_N
\end{bmatrix}
\end{equation}

Because an estimate of the camera matrix is needed for projection of the PCA vectors into the image plane, an iterative procedure is used where the camera is estimated, the shape parameters are solved for, and the process is repeated using the updated shape estimate to refine the camera estimate.

The optimization is performed on each image of the VGG-Face dataset~\cite{parkhi_15}.
Of the total 2.6 million images in the dataset, the estimation procedure converges successfully on approximately 2.4 million.
Figure~\ref{fig:vgg-face_estimated} shows qualitative results on a variety of test subjects.
In order to simplify processing and provide robustness to incorrect identity labels, the estimation process is run independently on each image, i.e.\ shape parameters are not optimized jointly across multiple images of a subject.

\section{Synthetic Pose Rendering}
Once the geometry of the head and the pose and internal parameters of the camera are estimated for a given facial image, novel poses of the face may be rendered by constructing a map from pixel coordinates of the rendered image to pixel coordinates of the original.
For each pixel in the rendered image, the corresponding 3-d ray given by the render camera model is intersected with the 3-d head model described by the estimated PCA coefficients.
The intersection point is then projected into the original image using the estimated camera model.
Due to self-occlusion, some parts of the head may not receive appearance information directly from the input image.
In these cases, the assumption of bilateral symmetry is used, and appearance information from the opposite side of the face (if available) is used to fill in the missing regions.
In addition to hard self-occlusion constraints, the angle at which the input camera ray intersects the viewed surface (relative to the local surface normal) is also taken into account; regions where the camera is viewing the surface ``head on'' are more reliable than those where the camera ray is grazing the surface at a shallow angle.
The angle of intersection is used to smooth out the occlusion map, creating a ``fade out'' where the surface curves away from the camera into occlusion.
This visibility score is used when weighting the combination of original appearance and symmetric appearance and creates a smooth transition from one to the other.

\begin{figure}[!ht]
  \includegraphics[width=0.5\textwidth]{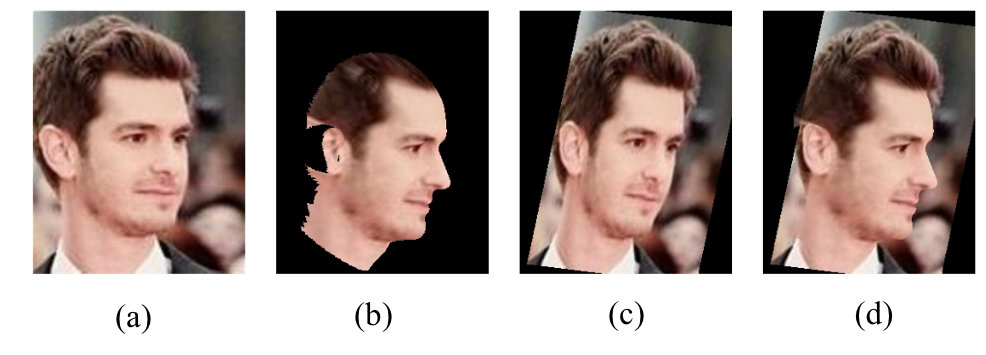}
  \caption{A simple planar warp is used to fill in the background information when rendering novel poses.
    From left to right: (a) Original image, (b) synthetic pose rendering, (c) planar background warp of original image, (d) synthetic pose with composited background.
  }
\label{fig:background_composite}
\end{figure}

\subsection{Background Composition}
\begin{figure*}[!ht]
  \centering
  \includegraphics[width=0.7\textwidth]{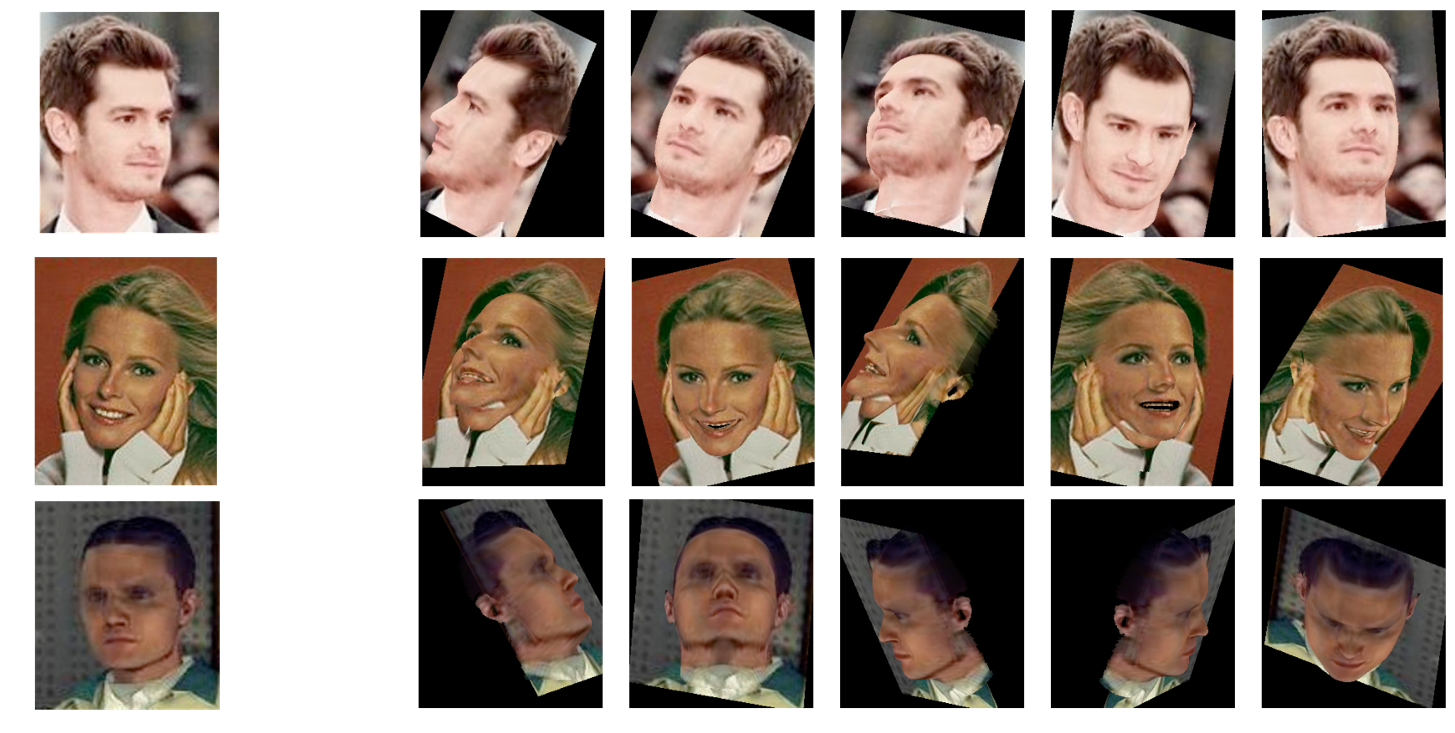}
  \caption{Results of random pose jittering on samples from the training set.  The original image is shown to the left.
  }
\label{fig:pose_results}
\end{figure*}

The mapping described above generates imagery for locations corresponding to the underlying 3-d head model, which does not include hair, body, or scene background.
This leads to unrealistic images as shown in Figure~\ref{fig:background_composite}(b).
To mitigate this problem, the 3-d rendering is composited onto a planar warp of the input image (Figure~\ref{fig:background_composite}(c)), resulting in a more natural looking image, as shown in Figure~\ref{fig:background_composite}(d).
The background warp is generated using a plane parallel to that of the input image, positioned such that the occluding contour of the head passes through the plane.
This facilitates as smooth a transition as possible from the 3-d rendering to the background.
Some representative results of pose jittering on the baseline VGG-Face training set are shown in Figure~\ref{fig:pose_results}.

\section{Synthetic Lighting Rendering}

In addition to synthetically altering the pose of a training image, the lighting conditions may also be modified.
Because the primary goal of the rendering is to generate large amounts of diverse imagery for training, a fast algorithm that makes use of three simplifying assumptions is used in place of a fully physically accurate representation: the face is assumed to exhibit Lambertian reflectance, the illumination in the input scene is assumed to be completely ambient (i.e.\ no directional component), and the illumination in the output scene is assumed to be a linear combination of the input ambient light and a single directional component.

\begin{figure}[h]
  \includegraphics[width=0.5\textwidth]{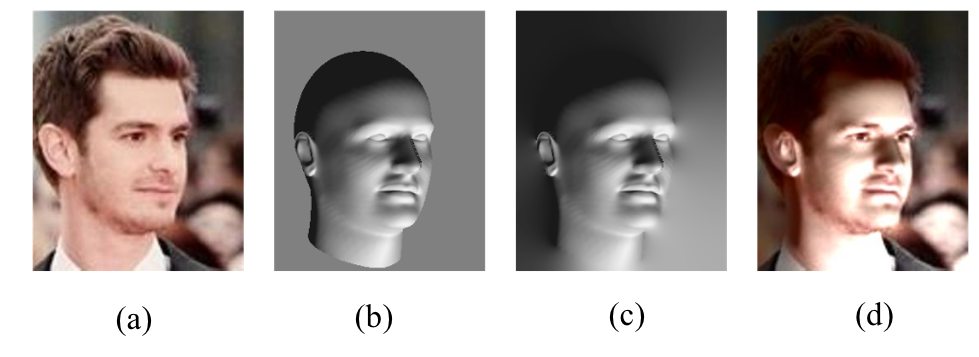}
  \caption{The 3-d surface normals of the estimated head model are used to approximate novel lighting conditions.
    From left to right: (a) Original image, (b) intensity modulation estimated using surface normals and desired lighting direction, (c) intensity modulation propagated to the full image, and (d) the output image.
  }
\label{fig:lighting_jitter_pipeline}
\end{figure}

\begin{figure*}[!ht]
  \centering
  \includegraphics[width=0.7\textwidth]{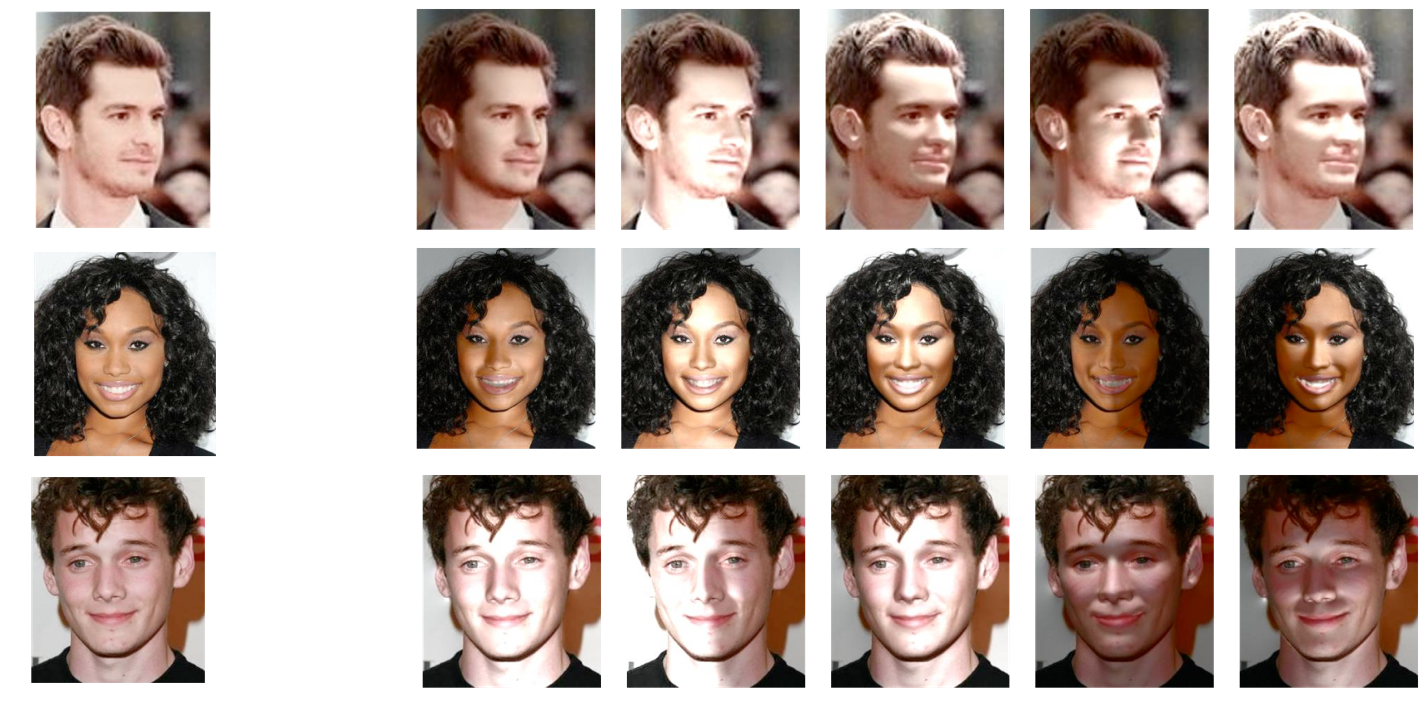}
  \caption{Results of random lighting jittering on samples from the training set.
The original image is shown to the left.
  }
\label{fig:relighting_results}
\end{figure*}

Given an estimated 3-d shape and camera model, the surface normals of the 3-d model are used to compute a modulation factor that determines the ratio of the reflected directional light to the reflected ambient light.
In order to avoid artifacts near the edge of the 3-d model, the modulation factors are extrapolated beyond its extent throughout the image by solving a discrete approximation of the Laplace Equation, using the known modulation values as boundary conditions.
The desired ambient and directional weights are then used to modulate the original image intensity (color hue and saturation are preserved) at each pixel location, ignoring self-shadowing effects.
Figure~\ref{fig:lighting_jitter_pipeline} provides a visualization of the relighting rendering steps described above.
Representative results obtained by randomly selecting directional lighting angle and intensity in the baseline training set are shown in Figure~\ref{fig:relighting_results}.

\section{Training Set Augmentation}
For each of the approximately 2.4 million images in the VGG-Face baseline training set for which the 3-d shape estimation successfully converges, five random pose jitters and five random lighting jitters are rendered.
The state of the art deep convolutional neural network presented by Parkhi et al.~\cite{parkhi_15} is used as a baseline for experimentation.
The network is trained using stochastic gradient descent for 20 epochs, with each epoch consisting of a full pass through the VGG-Face dataset in randomized order.
Two simple augmentation strategies are presented:
\begin{itemize}
 \item Random jitter: For each image used in training, a jittered version is used in place of the original with probability $P$.
 \item Dual jitter: For each image used in training, both the original image and a jittered version are used.
Note that this jittering method results in twice as many total images used over a fixed number of training epochs.
\end{itemize}
The jittered images are chosen with uniform probability from the available renderings in both cases.
In cases where the 3-d estimation or rendering failed to complete successfully, the original image is used.

\section{Results}
Due to the challenging variation in both pose and lighting across images of test subjects, evaluation is focused on the IJB-A datset~\cite{ijba}.
The IJB-A dataset features over 1.5 million images of 500 subjects, and evaluation protocols for both verification and identification.
The identification protocol reports the ability of an algorithm to identify an unknown subject (the ``probe'') within a list of known subjects (the ``gallery'').
One measure of identification performance is Rank-k retrieval, which is defined as the percentage of probes with the correct gallery mate in the top $k$ candidates returned by the algorithm's metric.
A CMC (cumulative match characteristic) curve depicts how Rank-k retrieval varies with respect to $k$.
In the case of an open test set (i.e.\ not all probe subjects exist in the gallery), it is useful to define a positive match for a given probe only if the maximum match score exceeds a given threshold.
This allows a user to set the match threshold according to an application-specific acceptable false positive rate, and receive only gallery matches that exceed this threshold.
The IJB-A protocol includes reporting of the True Positive Identification Rate (TPIR) at False Positive Identification Rates (FPIR) of $0.01$ and $0.1$.
The Decision Error Tradeoff (DET) curve measures the change in False Negative Identification Rate ($\text{FNIR} = 1 - \text{TPIR}$) with respect to FPIR\@.

The identification task is performed and evaluated using two distinct match scoring strategies: $L_2$ and ``Template Adaptation''.
The $L_2$ strategy uses of the output of the fc7 layer of the network presented by Parkhi et al.~\cite{parkhi_15} as a 4096-dimensional feature vector (with no triplet loss during training).
Euclidean distance between the (normalized) feature vectors is used as measure of dissimilarity under this regime.
Crosswhite et al.~\cite{crosswhite_16} presented a method of leveraging a large negative set of face encodings for improved identification performance over the $L_2$ metric, so-called ``Template Adaptation''.
Using this method, a linear ``one vs.\ all'' Support Vector Machine (SVM) is trained on each gallery and probe template, and the margins are used at test time as a measure of similarity.
In order to investigate the degree to which the gains achieved by Template Adaptation are complementary to any gains produced by training set augmentation, evaluation is performed under this regime as well.

\begin{figure*}[h!]
  \centering
  \begin{tabular}{cc}
    \multicolumn{2}{c}{\centering Pose Jittering}\\
    \includegraphics[width=0.4\textwidth]{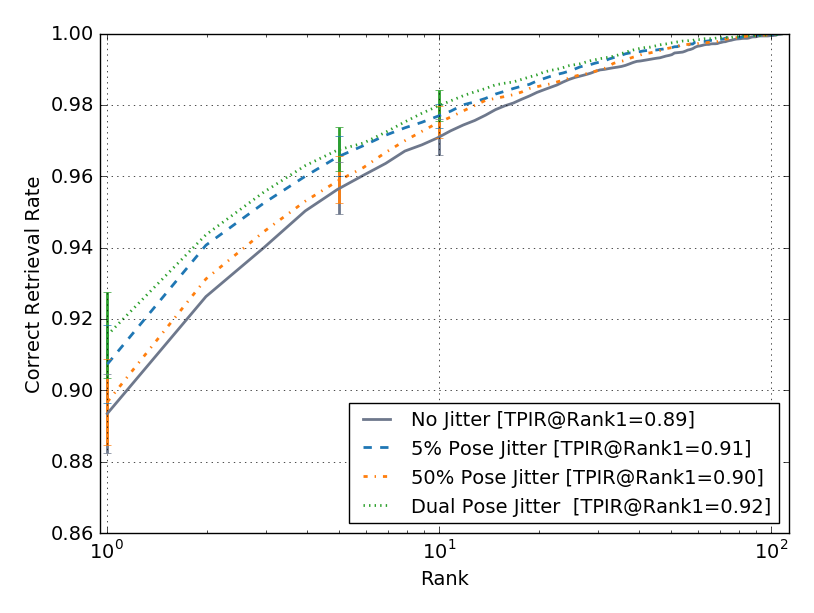} &
    \includegraphics[width=0.4\textwidth]{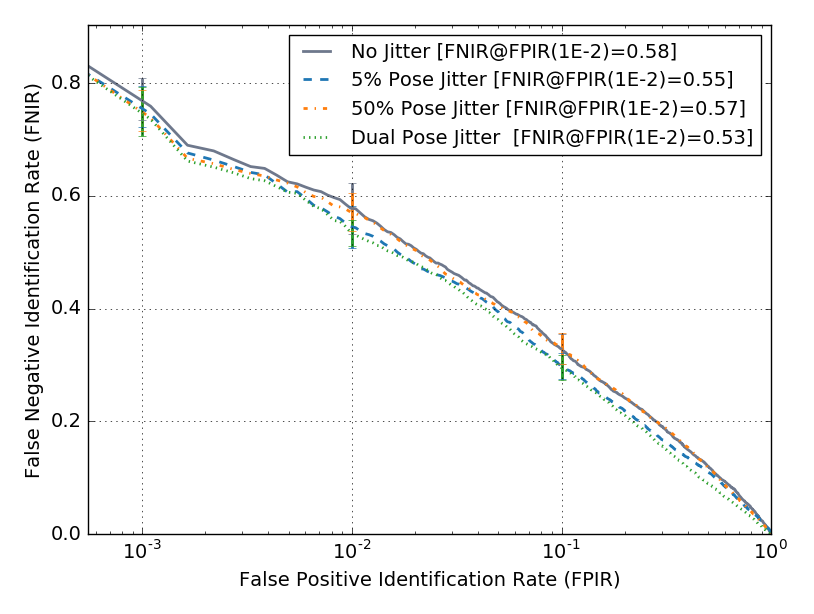} \\
    \multicolumn{2}{c}{Lighting Jittering }\\
    \includegraphics[width=0.4\textwidth]{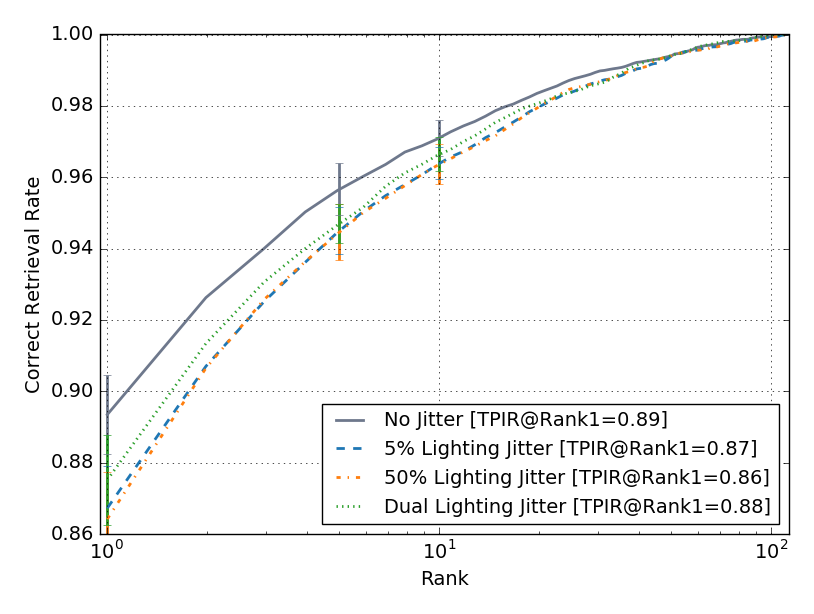} &
    \includegraphics[width=0.4\textwidth]{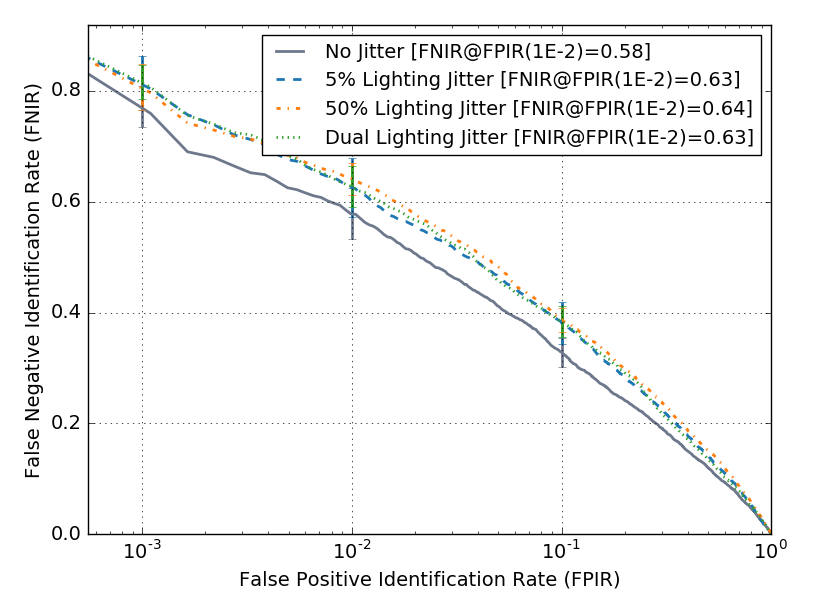} \\
    \multicolumn{2}{c}{Pose + Lighting Jittering }\\
  \includegraphics[width=0.4\textwidth]{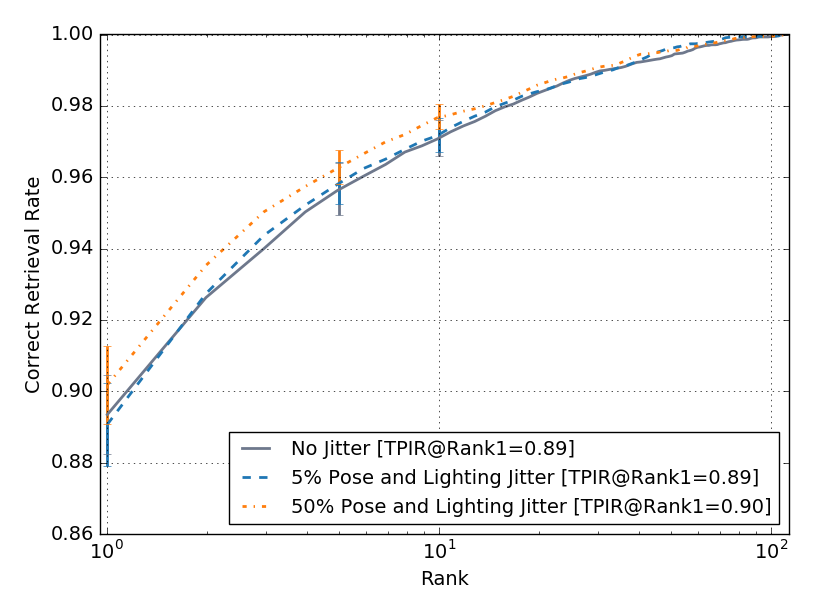}&
  \includegraphics[width=0.4\textwidth]{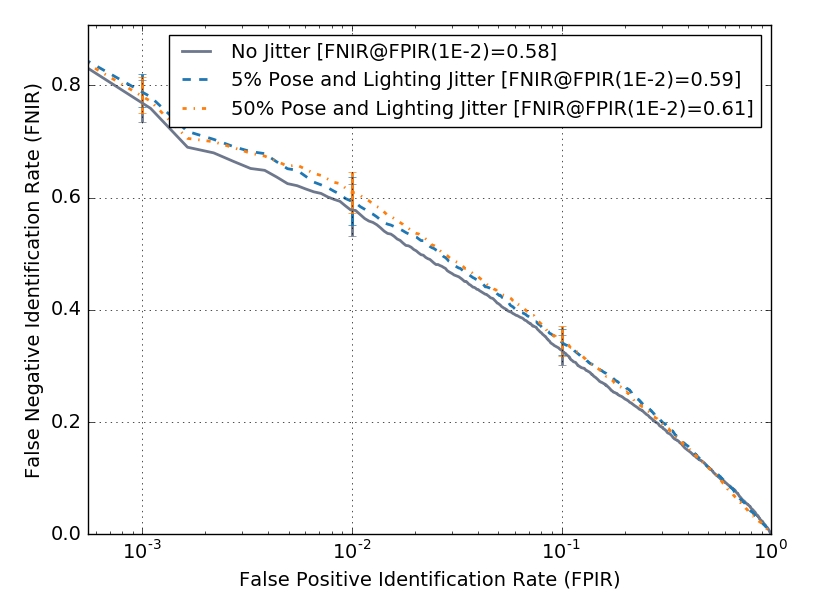}\\
\end{tabular}
\caption{$L_2$ metric CMC Results (left) and DET Results (right) for Pose Jittering (top), Lighting Jittering (middle) and both Pose and Lighting Jittering (bottom).
Error bars represent one standard deviation at IJB-A operating points.}
\label{fig:ijba_L2}
\end{figure*}

\begin{figure*}[h!]
  \centering
  \begin{tabular}{cc}
    \multicolumn{2}{c}{\centering Pose Jittering using Template Adaptation}\\
    \includegraphics[width=0.4\textwidth]{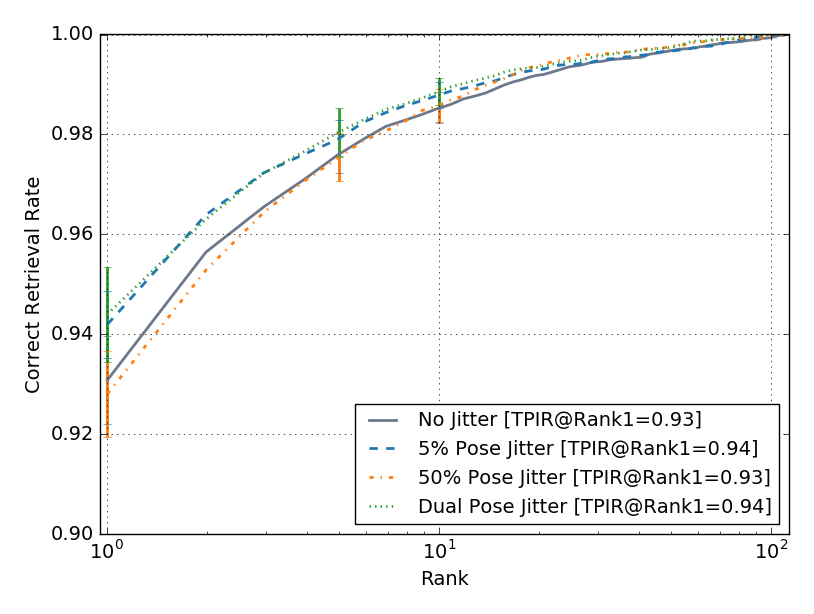} &
    \includegraphics[width=0.4\textwidth]{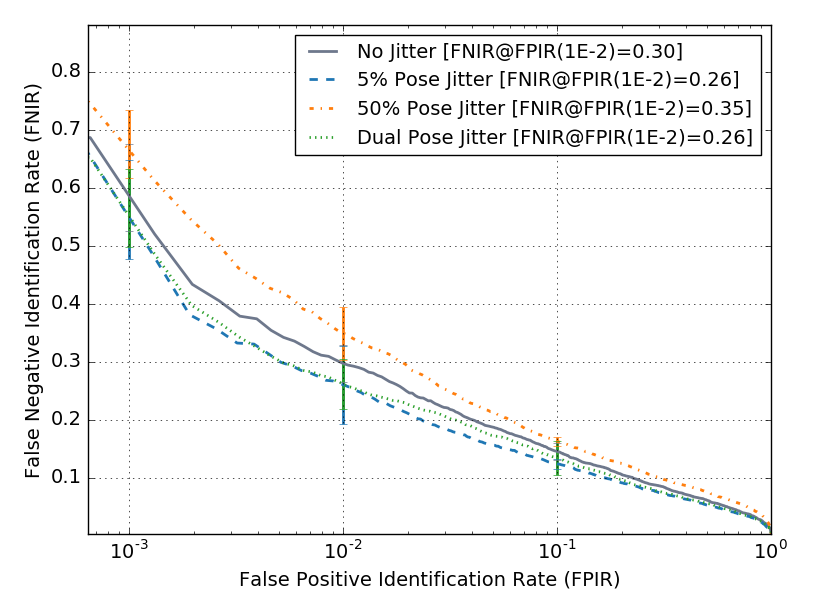} \\
    \multicolumn{2}{c}{\centering Lighting Jittering using Template Adaptation}\\
    \includegraphics[width=0.4\textwidth]{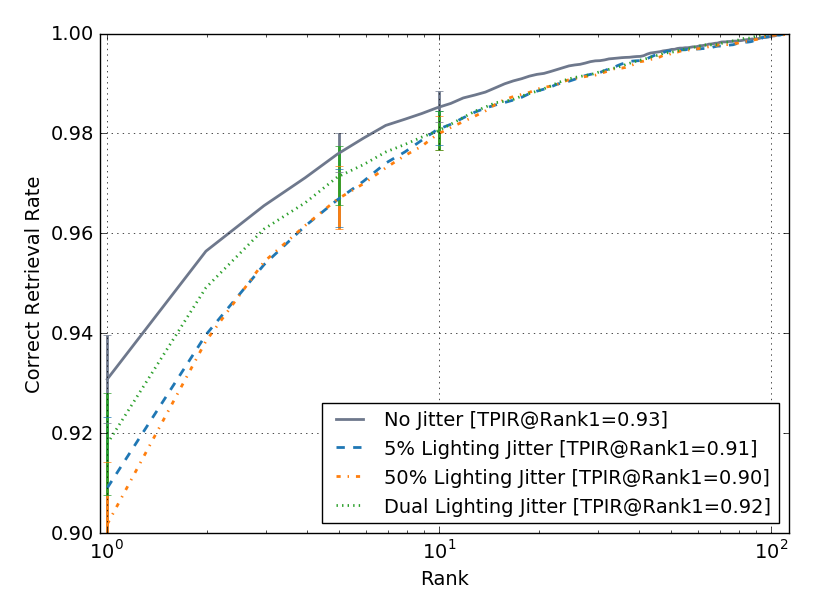} &
    \includegraphics[width=0.4\textwidth]{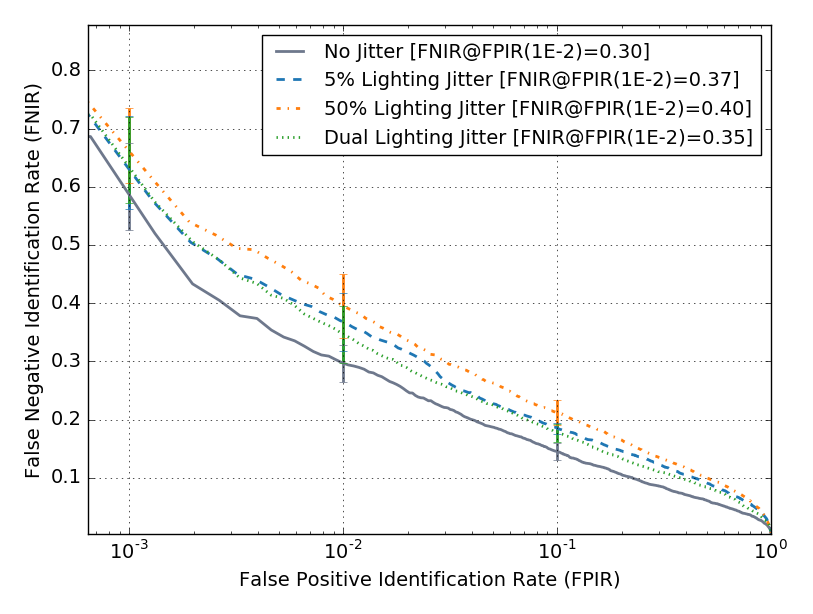} \\
    \multicolumn{2}{c}{\centering Pose + Lighting Jittering using Template Adaptation}\\
  \includegraphics[width=0.4\textwidth]{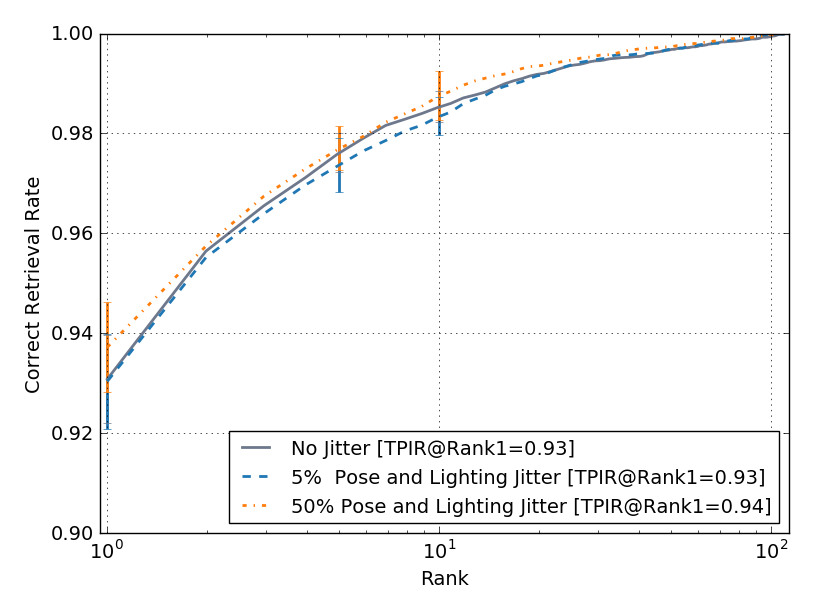}&
  \includegraphics[width=0.4\textwidth]{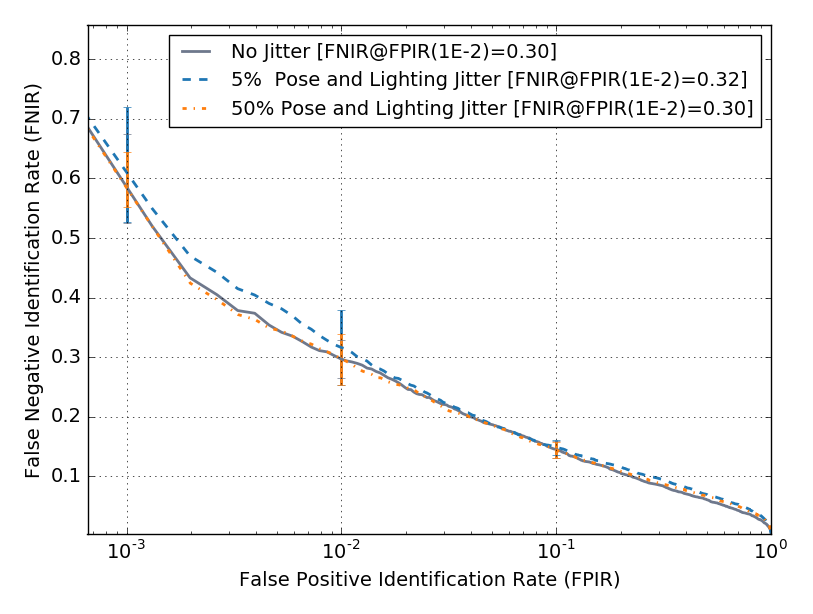}\\
\end{tabular}
\caption{Template Adaptation CMC Results (left) and DET Results (right) for Pose Jittering (top), Lighting Jittering (middle) and both Pose and Lighting Jittering (bottom).
Error bars represent one standard deviation at IJB-A operating points.}
\label{fig:ijba_TA}
\end{figure*}

IJB-A identification performance results for each of the augmented models are presented in Tables~\ref{tab:L2_results} and~\ref{tab:TA_results} under the $L_2$ metric and SVM Template Adaptation metric, respectively.
The general trend is that pose jittering provides a performance boost using both the $L_2$ metric and Template Adaptation, although the impact is slightly less in the latter case (but overall performance is better).
Conversely, lighting jittering appears to decrease identification performance in all reported cases.
The reasons for this discrepancy between pose jittering and lighting jittering performance is speculated upon in Section~\ref{sec:conclusion}.
Of the augmentation strategies presented, dual jittering appears most beneficial.
Random jittering at $P=0.05$ generally performs slightly worse than dual jittering, but requires half as many updates in the training process.
Interestingly, random jittering at $P=0.5$ degrades performance relative to the baseline results in all cases.
Finally, incorporating both pose and lighting jittering has a small positive effect on identification performance, but the benefit is smaller than that of applying pose jittering alone.
CMC and DET curves are shown in Figures~\ref{fig:ijba_L2} and~\ref{fig:ijba_TA} for both the $L_2$ and Template Adaptation match scoring strategies.

\begin{figure}[H]
  \includegraphics[width=0.4\textwidth]{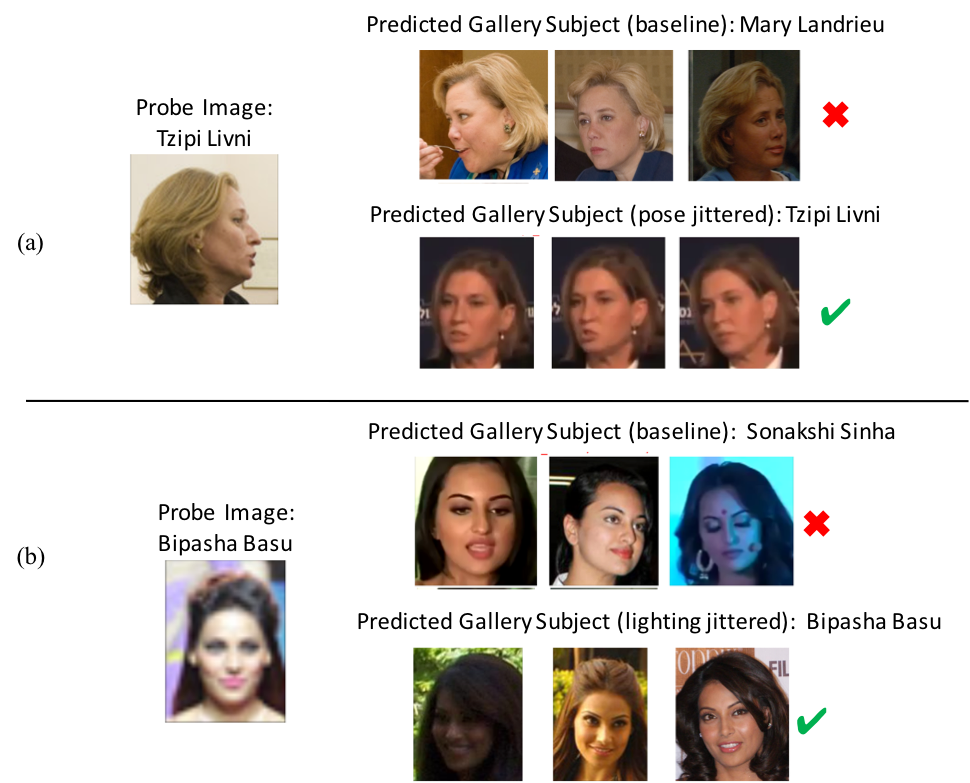}
  \caption{Probe images from the test set with the maximum matching error using the baseline network that are correctly matched using a network trained with synthetic augmentation.
(a) The network trained with pose jittering successfully matches a near-profile view to a gallery subject containing near-frontal views.
(b) The network trained with lighting jittering successfully matches the probe to the correct gallery subject despite the presence of harsh lighting conditions.}
\label{fig:id_examples}
\end{figure}

To understand the nature of the identification mistakes that the system is making, it can be useful to inspect individual failure cases.
Figure~\ref{fig:id_examples} shows the probe images from the IJB-A identification (split 1) test with maximal error according to the $L_2$ metric using the baseline network, conditioned on a correct match by the pose and lighting jittered (dual augmentation strategy) networks.
In other words, the probes represent the largest errors that are corrected using training set augmentation.
Note that although overall performance is degraded, the example shown in Figure~\ref{fig:id_examples}(b) does provide evidence that some measure of lighting invariance is obtained through augmentation.
Images within the gallery templates are ordered according to inverse $L_2$ distance from the mean gallery encoding in order to emphasize representative gallery images.

\section{Conclusion}
\label{sec:conclusion}
The proposed reconstruction, rendering, and relighting methods comprise an efficient and effective pipeline used to generate semi-synthetic imagery for the purpose of increasing lighting and pose variation in a face recognition training set.
Using a state of the art face recognition network and a benchmark specifically designed to test performance in unconstrained environments, quantitative evidence is presented that simple forms of training data pose jittering improves identification performance.
Similar forms of jittering applied to lighting conditions result in a decrease in recognition performance using the same metrics.
Further investigation is necessary to determine with certainty the source of this discrepancy, but a likely hypothesis is that severe lighting variations as added to the training set by the lighting jittering process do not actually reflect the statistics of the test set.
This is in contrast to large pose variation, which is an explicit goal of the IJB-A test set.
Anecdotal examples from the test set containing large pose and lighting variations support this hypothesis by demonstrating increased invariance to these conditions, albeit at the expense of lower performance under favorable lighting conditions.
Future work on the rendering pipeline will focus on both increased capability (e.g.\ expression jittering) and improved rendering performance (both speed and quality).
Additionally, further experimentation with augmentation strategies is warranted.
Specifically, augmentation strategies specifically designed to match expected statistics of pose and lighting variation in the test set will be investigated.
Such strategies may prove essential in specialized cases, such as those when pose is constrained by the collection environment.

\section*{Acknowledgement}
This research is based upon work supported by the Office of the Director of National Intelligence (ODNI), Intelligence Advanced Research Projects Activity (IARPA) under contract number 2014-14071600010.
The views and conclusions contained herein are those of the authors and should not be interpreted as necessarily representing the official policies or endorsements, either expressed or implied, of ODNI, IARPA, or the U.S. Government.
The U.S. Government is authorized to reproduce and distribute reprints for Governmental purpose notwithstanding any copyright annotation thereon.

\bibliography{effective_pose_augmentation}

\begin{thebibliography}{10}
\providecommand{\url}[1]{#1}
\csname url@samestyle\endcsname
\providecommand{\newblock}{\relax}
\providecommand{\bibinfo}[2]{#2}
\providecommand{\BIBentrySTDinterwordspacing}{\spaceskip=0pt\relax}
\providecommand{\BIBentryALTinterwordstretchfactor}{4}
\providecommand{\BIBentryALTinterwordspacing}{\spaceskip=\fontdimen2\font plus
\BIBentryALTinterwordstretchfactor\fontdimen3\font minus
  \fontdimen4\font\relax}
\providecommand{\BIBforeignlanguage}[2]{{%
\expandafter\ifx\csname l@#1\endcsname\relax
\typeout{** WARNING: IEEEtran.bst: No hyphenation pattern has been}%
\typeout{** loaded for the language `#1'. Using the pattern for}%
\typeout{** the default language instead.}%
\else
\language=\csname l@#1\endcsname
\fi
#2}}
\providecommand{\BIBdecl}{\relax}
\BIBdecl

\bibitem{blanz_99}
V.~Blanz and T.~Vetter, ``A morphable model for the synthesis of 3d faces,''
  \emph{Proceedings of the 26th annual conference on {Computer} graphics and
  interactive techniques}, pp. 187--194, 1999.

\bibitem{roth_15}
J.~Roth, Y.~Tong, and X.~Liu, ``Unconstrained 3d {Face} {Reconstruction},'' in
  \emph{CVPR}.\hskip 1em plus 0.5em minus 0.4em\relax IEEE, 2015, pp.
  2606--2615.

\bibitem{zhu_15}
X.~Zhu, Z.~Lei, J.~Yan, D.~Yi, and S.~Z. Li, ``High-{Fidelity} {Pose} and
  {Expression} {Normalization} for {Face} {Recognition} in the {Wild},'' in
  \emph{CVPR}, 2015, pp. 787--796.

\bibitem{gupta_10}
S.~Gupta, M.~K. Markey, and A.~C. Bovik, ``Anthropometric 3d {Face}
  {Recognition},'' \emph{International Journal of Computer Vision}, vol.~90,
  no.~3, pp. 331--349, Jun. 2010.

\bibitem{paysan_09}
P.~Paysan, R.~Knothe, B.~Amberg, S.~Romdhani, and T.~Vetter, ``A 3d face model
  for pose and illumination invariant face recognition,'' in \emph{Advanced
  {Video} and {Signal} {Based} {Surveillance}}.\hskip 1em plus 0.5em minus
  0.4em\relax IEEE, 2009, pp. 296--301.

\bibitem{taigman_13}
Y.~Taigman, M.~Yang, M.~Ranzato, and L.~Wolf, ``Deepface: {Closing} the gap to
  human-level performance in face verification,'' in \emph{CVPR}, 2013, pp.
  1701--1708.

\bibitem{chu_14}
B.~Chu, S.~Romdhani, and L.~Chen, ``3d-aided face recognition robust to
  expression and pose variations,'' in \emph{CVPR}.\hskip 1em plus 0.5em minus
  0.4em\relax IEEE, 2014, pp. 1907--1914.

\bibitem{masi_16}
I.~Masi, A.~T. Trần, T.~Hassner, J.~T. Leksut, and G.~Medioni, ``Do {We}
  {Really} {Need} to {Collect} {Millions} of {Faces} for {Effective} {Face}
  {Recognition}?'' in \emph{ECCV}.\hskip 1em plus 0.5em minus 0.4em\relax
  Springer International Publishing, Oct. 2016, no. 9909, pp. 579--596.

\bibitem{kazemi_14}
V.~Kazemi and J.~Sullivan, ``One {Millisecond} {Face} {Alignment} with an
  {Ensemble} of {Regression} {Trees},'' in \emph{CVPR}.\hskip 1em plus 0.5em
  minus 0.4em\relax IEEE, 2014, pp. 1867--1874.

\bibitem{dlib_09}
D.~E. King, ``Dlib-ml: A machine learning toolkit,'' \emph{Journal of Machine
  Learning Research}, vol.~10, pp. 1755--1758, 2009.

\bibitem{hartley_zisserman}
R.~Hartley and A.~Zisserman, \emph{Multiple {View} {Geometry} in {Computer}
  {Vision}}, 2nd~ed.\hskip 1em plus 0.5em minus 0.4em\relax Cambridge
  University Press, 2004.

\bibitem{facegen}
{Singular Inversions}, ``Facegen,'' \url{http://www.facegen.com/}, 2016.

\bibitem{parkhi_15}
O.~M. Parkhi, A.~Vedaldi, and A.~Zisserman, ``Deep face recognition,''
  \emph{BMVC}, vol.~1, no.~3, p.~6, 2015.

\bibitem{ijba}
B.~Klare, B.~Klein, E.~Taborsky, A.~Blanton, J.~Cheney, K.~Allen, P.~Grother,
  A.~Mah, M.~Burge, and A.~Jain, ``Pushing the frontiers of unconstrained face
  detection and recognition: {IARPA} {Janus} {Benchmark} {A},'' in
  \emph{CVPR}.\hskip 1em plus 0.5em minus 0.4em\relax IEEE, Jun. 2015, pp.
  1931--1939.

\bibitem{crosswhite_16}
N.~Crosswhite, J.~Byrne, O.~M. Parkhi, C.~Stauffer, Q.~Cao, and A.~Zisserman,
  ``Template {Adaptation} for {Face} {Verification} and {Identification},''
  \emph{arXiv:1603.03958 [cs]}, Mar. 2016, arXiv: 1603.03958.

\end{thebibliography}
\bibliographystyle{IEEEtran}

\clearpage


\begin{table*}
  \centering
  \small
  \tabcolsep=0.11cm
  \begin{tabular}{|c|ccccc|}
    \hlineB{2.5}
    Algorithm & TPIR @ FPIR = 1e-1 & TPIR @ FPIR = 1e-2 & Rank-1 & Rank-5 & Rank-10\\
    \hlineB{2.5}
    No Jitter & 0.660 $\pm$ 0.027 & 0.413 $\pm$ 0.045 & 0.893 $\pm$ 0.011 & 0.957 $\pm$ 0.007 & 0.971 $\pm$ 0.005\\
    \hline
5\% Pose Jitter & 0.692 $\pm$ 0.024 & 0.448 $\pm$ 0.039 & 0.907 $\pm$ 0.011 & 0.966 $\pm$ 0.006 & 0.977 $\pm$ 0.003\\
50\% Pose Jitter & 0.660 $\pm$ 0.028 & 0.421 $\pm$ 0.034 & 0.897 $\pm$ 0.012 & 0.959 $\pm$ 0.007 & 0.975 $\pm$ 0.005\\
Dual Pose Jitter  & \textbf{0.696 $\pm$ 0.022} & \textbf{0.460 $\pm$ 0.024} & \textbf{0.915 $\pm$ 0.012} & \textbf{0.968 $\pm$ 0.006} & \textbf{0.980 $\pm$ 0.004}\\
    \hline
5\% Light Jitter & 0.604 $\pm$ 0.040 & 0.366 $\pm$ 0.053 & 0.867 $\pm$ 0.012 & 0.945 $\pm$ 0.007 & 0.964 $\pm$ 0.004\\
50\% Light Jitter & 0.599 $\pm$ 0.024 & 0.352 $\pm$ 0.030 & 0.864 $\pm$ 0.013 & 0.945 $\pm$ 0.008 & 0.964 $\pm$ 0.006\\
Dual Light Jitter & 0.603 $\pm$ 0.028 & 0.364 $\pm$ 0.037 & 0.875 $\pm$ 0.013 & 0.947 $\pm$ 0.006 & 0.967 $\pm$ 0.005\\
    \hline
5\% Pose and Light Jitter & 0.645 $\pm$ 0.025 & 0.399 $\pm$ 0.042 & 0.891 $\pm$ 0.012 & 0.958 $\pm$ 0.006 & 0.972 $\pm$ 0.005\\
50\% Pose and Light Jitter & 0.645 $\pm$ 0.027 & 0.384 $\pm$ 0.038 & 0.902 $\pm$ 0.011 & 0.963 $\pm$ 0.005 & 0.977 $\pm$ 0.004\\
    \hline
  \end{tabular}
  \caption{Quantitative results for the IJB-A protocol under the $L_2$ metric }
\label{tab:L2_results}
\end{table*}


\begin{table*}
  \centering
  \small
  \tabcolsep=0.11cm
  \begin{tabular}{|c|ccccc|}
    \hlineB{2.5}
    Algorithm & TPIR @ FPIR = 1e-1 & TPIR @ FPIR = 1e-2 & Rank-1 & Rank-5 & Rank-10\\
    \hlineB{2.5}
    No Jitter & 0.848 $\pm$ 0.015 & 0.697 $\pm$ 0.030 & 0.931 $\pm$ 0.009 & 0.976 $\pm$ 0.004 & 0.985 $\pm$ 0.003 \\
    \hline
    5\% Pose Jitter & \textbf{0.870 $\pm$ 0.009} & \textbf{0.734 $\pm$ 0.068} & 0.942 $\pm$ 0.007 & 0.979 $\pm$ 0.004 & 0.988 $\pm$ 0.003 \\
50\% Pose Jitter & 0.832 $\pm$ 0.008 & 0.648 $\pm$ 0.047 & 0.928 $\pm$ 0.009 & 0.976 $\pm$ 0.005 & 0.986 $\pm$ 0.003 \\
Dual Pose Jitter & 0.860 $\pm$ 0.029 & \textbf{0.734 $\pm$ 0.042} & \textbf{0.944 $\pm$ 0.009} & \textbf{0.981 $\pm$ 0.005} & \textbf{0.989 $\pm$ 0.003} \\
\hline
5\% Light Jitter & 0.805 $\pm$ 0.011 & 0.626 $\pm$ 0.050 & 0.909 $\pm$ 0.014 & 0.967 $\pm$ 0.006 & 0.981 $\pm$ 0.003 \\
50\% Light Jitter & 0.779 $\pm$ 0.022 & 0.599 $\pm$ 0.055 & 0.902 $\pm$ 0.012 & 0.967 $\pm$ 0.006 & 0.980 $\pm$ 0.003 \\
Dual Light Jitter & 0.813 $\pm$ 0.017 & 0.643 $\pm$ 0.049 & 0.918 $\pm$ 0.010 & 0.972 $\pm$ 0.006 & 0.981 $\pm$ 0.004 \\
\hline
5\%  Pose and Light Jitter & 0.844 $\pm$ 0.015 & 0.678 $\pm$ 0.063 & 0.930 $\pm$ 0.010 & 0.974 $\pm$ 0.005 & 0.983 $\pm$ 0.004 \\
50\% Pose and Light Jitter & 0.850 $\pm$ 0.013 & 0.698 $\pm$ 0.043 & 0.937 $\pm$ 0.009 & 0.977 $\pm$ 0.004 & 0.988 $\pm$ 0.005 \\
    \hline
  \end{tabular}
  \caption{Quantitative results for the IJB-A protocol under using Template Adaptation metric.}
\label{tab:TA_results}
\end{table*}

\end{document}